\documentclass[11pt,a4paper]{article}
\usepackage[hyperref]{acl2019}
\usepackage{times}
\usepackage{latexsym}

\usepackage{url}
\usepackage{array}

\usepackage{xcolor}
\usepackage{multirow}
\usepackage{graphicx}

\aclfinalcopy 
\def\aclpaperid{151} 

\newcolumntype{L}[1]{>{\raggedright\let\newline\\\arraybackslash\hspace{0pt}}m{#1}}

\title{Enriching Neural Models with Targeted Features for Dementia Detection}

\author{Flavio Di Palo \and
  Natalie Parde \\
  Department of Computer Science \\
  University of Illinois at Chicago \\
  \texttt{\{fdipal2, parde\}@uic.edu} \\}

\date{}

\begin{document}
\maketitle
\begin{abstract}
Alzheimer's disease (AD) is an irreversible brain disease that can dramatically reduce quality of life, most commonly manifesting in older adults and eventually leading to the need for full-time care. Early detection is fundamental to slowing its progression; however, diagnosis can be expensive, time-consuming, and invasive.  In this work we develop a neural model based on a CNN-LSTM architecture that learns to detect AD and related dementias using targeted and implicitly-learned features from conversational transcripts.  Our approach establishes the new state of the art on the DementiaBank dataset, achieving an F1 score of 0.929 when classifying participants into AD and control groups.

\end{abstract}

\section{Introduction}

Older adults constitute a growing subset of the population.  In the United States, adults over age 65 are expected to comprise one-fifth of the population by 2030, and a larger proportion of the population than those under 18 by 2035 \cite{us_population_projection}.  In Japan---perhaps the most extreme example of shifting age demographics---42.4\% of the population is expected to be aged 60 or over by 2050 \cite{global_population_projection}.  This will necessitate that age-related physical and cognitive health issues become a foremost concern not only because they will impact such a large population, but because there will be a proportionally smaller number of human caregivers available to diagnose, monitor, and remediate those conditions.  Artificial intelligence offers the potential to fill many of these deficits, and already, elder-focused research is underway to test intelligent systems that monitor and assist with activities of daily living \cite{Lotfi2012}, support mental health \cite{paro}, promote physical well-being \cite{sarma-etal-2014-framework}, and encourage cognitive exercise \cite{parde_nielsen_aied}.  

Perhaps some of the most pressing issues beleaguering an aging population are Alzheimer's disease (AD) and other age-related dementias.  Our interest lies in fostering early diagnosis of these conditions.  Although there are currently no cures, with early diagnosis the symptoms can be managed and their impact on quality of life may be minimal.  However, there can be many barriers to early diagnosis, including cost, location, mobility, and time.

Here, we present preliminary work towards automatically detecting whether individuals suffer from AD using only conversational transcripts.  This solution addresses the above barriers by providing a diagnosis technique that could eventually be employed free of cost and in the comfort of one's home, at whatever time works best.  Our contributions are as follows:
\begin{enumerate}
    \item We introduce a hybrid Convolutional Neural Network (CNN) and Long Short Term Memory Network (LSTM) approach to dementia detection that takes advantage of both targeted and implicitly learned features to perform classification.
    \item We explore the effects of a bi-directional LSTM and attention mechanism on both our model and the current state-of-the-art for dementia detection. 
    \item We empirically demonstrate that our technique outperforms the current state of the art, and suggest directions for future work that we expect to further improve performance.
\end{enumerate}

\section{Related Work}

The task of automatically detecting dementia in conversational transcripts is not new. In \newcite{Fraser2016LinguisticFI}, the authors tackled the task using features associated with many linguistic phenomena, including part of speech tags, syntactic complexity, psycholinguistic characteristics, vocabulary richness, and many others. They trained a logistic regression model to distinguish between dementia-affected and healthy patients, achieving an accuracy of 81\%.  In our work here we consider some of the features found to be informative in this work; in particular, psycholinguistic features.

\newcite{Guinn} studied Alzheimer's-related dementia (AD) specifically. The authors selected 14 linguistic features to perform syntactic, semantic, and disfluency modeling.  In doing so, they checked for the presence of filler words, repetitions, and incomplete words, and additionally incorporated counts indicating the number of syllables used per minute.
Using this feature set, the authors trained a decision tree classifier to make predictions for 80 conversational samples from 31 AD and 57 non-AD patients. Their model achieved an accuracy of 79.5\%.

\newcite{Orimaye2014} considered syntactic features, computed from syntactic tree structures, and various lexical features to evaluate four machine learning algorithms for dementia detection.  The algorithms considered included a decision tree, na\"{i}ve Bayes, SVM with a radial basis kernel, and a neural network. On a dataset containing 242 AD and 242 healthy individuals, they found that compared to other algorithms, SVM exhibited the best performance with an accuracy score of 74\%, a recall of 73\%, and a precision of 75\%.

\newcite{Yancheva2016VectorspaceTM} used automatically-generated topic models to extract a small number of semantic features (12), which they then used to train a random forest classifier.  Their approach achieved an F1 Score of 0.74 in binary classification of control patients versus dementia-affected patients. This is comparable to results (F1 Score=0.72) obtained with a much larger set of lexicosyntactic and acoustic features. Ultimately, \citeauthor{Yancheva2016VectorspaceTM} found that combining these varied feature types improved their F1 Score to 0.80. 

Finally, \newcite{karlek2018} proposed a CNN-LSTM neural language model and explored the effectiveness of part-of-speech (POS) tagging the conversational transcript to improve  classification accuracy for AD patients. They divided patient interviews into single utterances, and rather than classifying at the patient level, they made their predictions at the utterance level. Their model achieved an accuracy of 91\%.  Unfortunately, the dataset on which their classifier was trained is imbalanced, and no other performance metrics were reported.  This makes it difficult to fully understand the capabilities of their model.  Here, in addition to our other contributions, we extend their work by considering a full-interview classification scenario and providing more detailed classification metrics to assess the classifier's quality.

\section{Data}

We use a subset of DementiaBank \cite{becker1994natural} for our work here.  DementiaBank is a dataset gathered as part of a protocol administered by the Alzheimer and Related Dementias Study at the University of Pittsburgh School of Medicine. It contains spontaneous speech from individuals who do (AD group) and do not (control group) present different kinds of dementia.  Participants in the dataset performed several different tasks:
\begin{itemize}
    \item \textbf{Fluency: } Participants were asked to name words belonging to a given category or that start with a given letter.

    \item \textbf{Recall: } Participants were asked to recall a story from their past experience.

    \item \textbf{Sentence: } Participants were asked to construct a simple sentence with a given word, or were asked if a given sentence made sense.
    
    \item \textbf{Cookie Theft: } Participants were asked to verbally describe an eventful image illustrating, among other elements, a child attempting to steal a cookie.  For this task, the participant's and interviewer's speech utterances were recorded and manually transcribed according to the TalkBank CHAT protocol \cite{talkbank}.

\end{itemize}

Of these tasks, Cookie Theft provides the largest source of unstructured text.  Thus, it is the data subset that we use for our work here.  In total, the Cookie Theft sub-corpus consists of 1049 transcripts from 208 patients suffering from dementia (AD group) and 243 transcripts from 104 healthy elderly patients (control, or CT, group), for a total of 1229 transcripts.  Dataset statistics are provided in Table \ref{tab:dataset_stats}.  For each participant, DementiaBank also provides demographic information including age, gender, education, and race.  We use all available transcripts, and randomly separate them into 81\% training, 9\% validation, and 10\% testing.

\begin{table}[t]
\small
\centering
\renewcommand{\arraystretch}{1.2}
\begin{tabular}{lccc}
\hline
\multicolumn{1}{|c}{\textbf{}} & \multicolumn{1}{c}{\textbf{Total}} & \textbf{AD} & \multicolumn{1}{c|}{\textbf{CT}} \\ \hline
\textbf{Number of Participants}                      & 312                                 & 208               & 104              \\ 
\textbf{Number of Transcripts}                   & 1229                                & 1049              & 243              \\ 
\textbf{Median Interview Length}          & 73                                  & 65                & 97               \\ 
\end{tabular}
\caption{Dataset statistics including the number of participants, the number of transcripts, and the median interview length.
Interview length is computed as the number of words spoken by the patient during the interview.}
\label{tab:dataset_stats}
\end{table}

\section{Methods}
We propose a neural network architecture designed to classify patients into the two groups mentioned previously: those suffering from dementia, and those who are not.  The architecture takes as input transcriptions of the patients' spoken conversations.  The transcripts are of moderate length (the average participant spoke 73 words across 16.8 utterances). We consider all participant speech in a single block rather than splitting the interview into separate utterances, allowing the model to consider the entire interview context in a manner similar to a real diagnosis scenario.

\subsection{Model Architecture}

The model architecture proposed is a CNN-LSTM \cite{CNNLSTMZhou2015ACN} with several modifications:
\begin{itemize}

    \item We introduced a dense neural network at the end of the LSTM layer to also take into consideration linguistic features that have been considered significant by previous research \cite{karlek2018,Tomsals}.
    \item Rather than a classic unidirectional LSTM, we used a bi-directional LSTM and inserted an attention mechanism on the hidden states of the LSTM. In this way we expect our model to identify specific linguistic patterns related to dementia detection. In addition, the attention mechanism has proven to lead to performance improvements when long sequences are considered \cite{Yang2016HierarchicalAN}.
    \item We added class weights to the loss function during training to take into account the dataset imbalance. 
    
\end{itemize}

\begin{figure}
\centering
  \includegraphics[width=.7\linewidth]{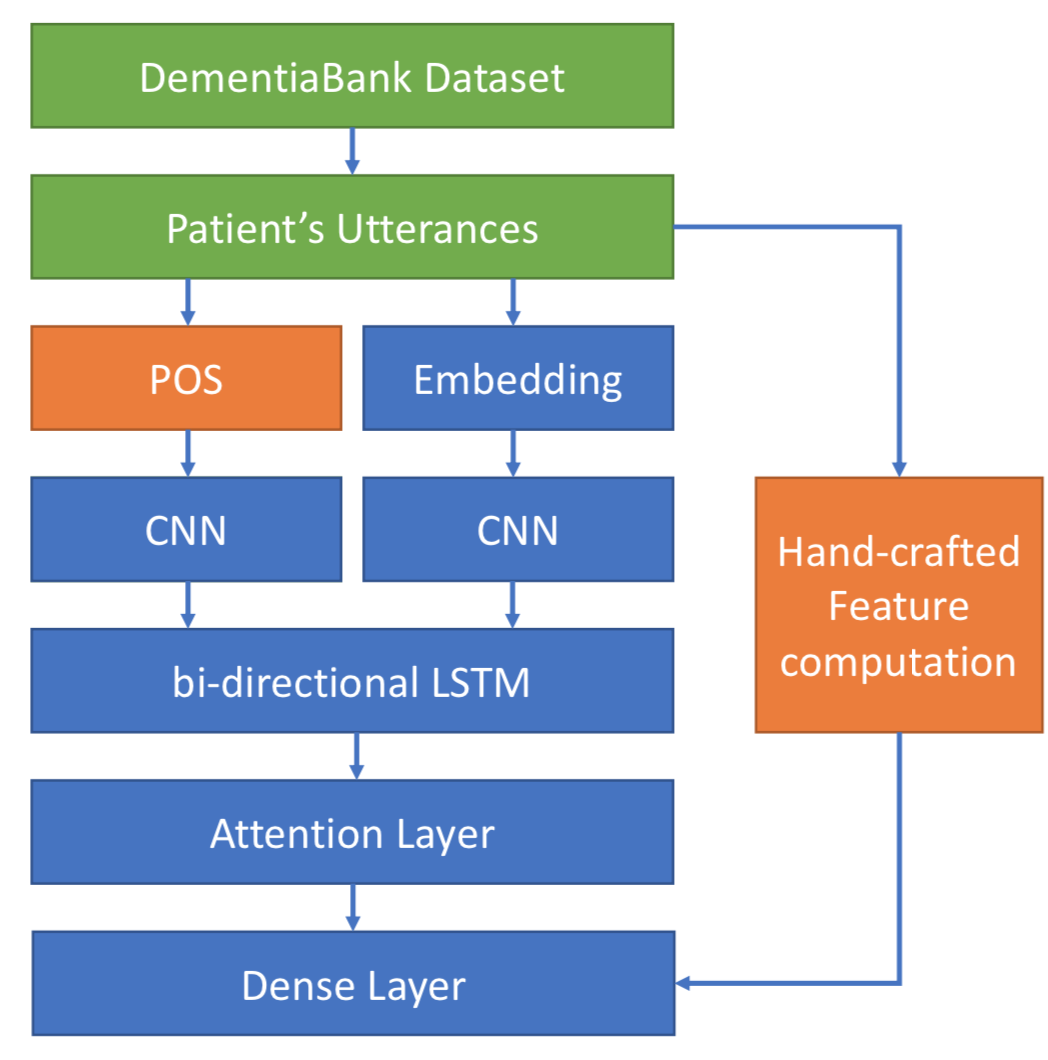}
  \caption{Model architecture.}
  \label{fig:model}
\end{figure}

We illustrate the architecture in Figure \ref{fig:model}.  We preprocess each full interview transcript from DementiaBank by removing interviewer utterances and truncating the length of the remaining text to 73 words.  This is done so that (a) each instance is of a uniform text size, and (b) the instances are of relatively substantial length, thereby providing adequate material with which to assess the health of the patient.  Seventy-three words represents the median (participant-only) interview length; thus, 50\% of instances include the full interview (padded as needed), and 50\% of instances are truncated to their first 73 words.  The interviews are tokenized into single word tokens, and POS tags\footnote{We compute POS tags using NLTK (\url{https://www.nltk.org/}).} are computed for each token.

The model takes two inputs: the tokenized interview, and the corresponding POS tag list. Word embeddings for the interview text tokens are computed using pre-trained 300 dimensional GloVe embeddings trained on the Wikipedia 2014 and Gigaword 5 dataset \cite{pennington2014glove}. The POS tag for each word is represented as a one-hot-encoded vector. The word embeddings and POS vectors are input to two different CNNs utilizing the same architecture, and the output of the two CNNs is then flattened and given as input to a bi-directional LSTM with an attention mechanism. 

The output of the bi-directional LSTM is then given as input to a dense neural network which also takes into consideration linguistic features that have proven to be effective in previous literature, as well as some demographic features (see further discussion in Section \ref{handcrafted_features}). The final outcome of the model is obtained with a single neuron at the end of the dense layer having a sigmoid activation function.  We implement the model using Keras.\footnote{\url{https://keras.io/}}  The advantage of our hybrid architecture that considers both implicitly-learned and engineered features is that it can jointly incorporate information that may be useful but latent to the human observer and information that directly encodes findings from clinical and psycholinguistic literature.

\subsection{Targeted Features}
\label{handcrafted_features}
Previous research has shown the effectiveness of neural models trained on conversational transcripts at identifying useful features for dementia classification \cite{review,karlek2018,Orimaye2018}. Nevertheless, other information that has proven to be crucial to the task cannot be derived from interview transcripts themselves. 
Inspired by \citeauthor{karlek2018}'s \shortcite{karlek2018} finding that adding POS tags as features improved the performance of their neural model, we sought to enrich our model with other engineered features that have proven effective in prior dementia detection work.  We describe those features Table \ref{tab:features}.  

Each of the token-level (psycholinguistic or sentiment) features was averaged across all tokens in the instance, allowing us to obtain a participant-level feature vector to be coupled with the participant-level demographic features.  These features were then concatenated with the output of our model's attention layer and the resulting vector was given as input to a dense portion of the neural network that performed the final classification. Sentiment scores were obtained using NLTK's sentiment library and psycholinguistic scores were obtained from  an open source repository\footnote{\url{https://github.com/vmasrani/dementia\_classifier}} based on the work of \newcite{Fraser2016LinguisticFI}.  As noted earlier, demographic information was included with the DementiaBank dataset.

\begin{table}[t]
\small
\centering
\renewcommand{\arraystretch}{1.2}
\begin{tabular}{|lL{1.75cm}m{3.5cm}}
\hline
 & \textbf{Feature} & \multicolumn{1}{l|}{\textbf{Description}} \\ \hline
\multicolumn{1}{l}{\multirow{4}{*}{\textbf{Psych.}}} & \textit{Age of Acquisition} & The age at which a particular word is usually learned. \\ 
\multicolumn{1}{l}{} & \textit{Concreteness} & A measure of a word's tangibility. \\ 
\multicolumn{1}{l}{} & \textit{Familiarity} & A measure of how often one might expect to encounter a word. \\ 
\multicolumn{1}{l}{} & \textit{Imageability} & A measure of how easily a word can be visualized. \\ \hline
\multicolumn{1}{l}{\textbf{Sent.}} & \textit{Sentiment} & A measure of a word's sentiment polarity. \\ \hline
\multicolumn{1}{l}{\multirow{2}{*}{\textbf{Demo.}}} & \textit{Age} & The participant's age at the time of the visit. \\ 
\multicolumn{1}{l}{} & \textit{Gender} & The participant's gender. \\ 
\end{tabular}
\caption{Targeted psycholinguistic, sentiment, and demographic features considered by the model.}
\label{tab:features}
\end{table}

    



\subsection{Class Weight Correction}
Since the DementiaBank dataset is unbalanced (more participants suffer from dementia than not), we noticed that even when high accuracy was achieved by previously proposed models, they resulted in poor precision scores.  This was because those classifiers were prone to producing false positive outcomes. To combat this issue, we tuned the loss function of our model such that it more severely penalized misclassifying the less frequent class.

\section{Evaluation}

\subsection{Baseline Approach}
We selected the \textsc{C-LSTM} model developed by \newcite{karlek2018} as our baseline approach.  This model represents the current state of the art for dementia detection on the DementiaBank dataset \cite{review}.

\subsection{Experimental Setup}

\begin{table*}[t]
\small
\centering
\renewcommand{\arraystretch}{1.2}
\begin{tabular}{lccccccccc}
\hline
\multicolumn{1}{|l}{\textbf{Approach}} & \textbf{Accuracy} & \textbf{Precision} & \multicolumn{1}{l}{\textbf{Recall}} & \textbf{F1}       & \textbf{AUC}      & \textbf{TN} & \textbf{FP} & \textbf{FN} & \multicolumn{1}{c|}{\textbf{TP}} \\ \hline
\textsc{C-LSTM}                               & 0.8384          & 0.8683           & 0.9497                            & 0.9058          & 0.9057          & 6.3         & 15.6        & 5.3         & 102.6                            \\
\textsc{C-LSTM-Att}                     & 0.8333          & 0.8446           & 0.9778                            & 0.9061          & 0.9126          & 2.6         & 19.3        & 2.3         & 105.6                            \\
\textsc{C-LSTM-Att-w}                     & 0.8512          & 0.9232           & 0.8949                            & 0.9084          & 0.9139          & 14.0        & 8.0         & 11.3        & 96.6                             \\
\textsc{OURS}                       & 0.8495          & 0.8508           & \textbf{0.9965}                   & 0.9178          & 0.9207          & 1.0         & 16.6        & 0.3         & 95.0                             \\
\textsc{OURS-Att}                      & 0.8466          & 0.8525           & 0.9895                            & 0.9158          & \textbf{0.9503} & 1.3         & 16.3        & 1.0         & 94.3                             \\
\textsc{OURS-Att-w}               & \textbf{0.8820} & \textbf{0.9312}  & 0.9298                            & \textbf{0.9305} & 0.9498          & 11.0        & 6.6         & 6.6         & 88.6   

\end{tabular}
\caption{Performance of evaluated models.}
\label{tab:performance}
\end{table*}

We split the dataset into 81\% training, 9\% validation, and 10\% testing.  Each data sample represents a patient interview and its associated demographic characteristics. In order to have a more robust evaluation, we split the dataset multiple times.  Thus, each model has been trained, validated, and tested using three different random shufflings of the data with different random seeds. The results presented are the average of the results that each model achieved over the three test sets. 

To measure performance we consider Accuracy, Precision, Recall, F1 Score, Area Under the Curve (AUC), and the number of True Negative (TN), False Positive (FP), False Negative (FN), and True Positive (TP) classifications achieved by each approach on the test set.  All metrics except AUC used a classification threshold of 0.5.

We compared six different models on the described task: two main architectures (ours and the state of the art approach developed by \newcite{karlek2018}), each with several variations.  The baseline version of \citeauthor{karlek2018}'s \shortcite{karlek2018} model (\textsc{C-LSTM}) is used directly, without any modification. Our architecture is \textsc{OURS}. 
For both architectures we then consider the effects of switching to a bidirectional LSTM and adding an attention mechanism (\textsc{-Att}) and the effects of class weight correction inside the loss function (\textsc{-w}).

\subsection{Results}
We report performance metrics for each model in Table \ref{tab:performance}.  As is demonstrated, our proposed model achieves the highest performance in Accuracy, Precision, Recall, F1, and AUC.  It outperforms the state of the art (\textsc{C-LSTM}) by 5.2\%, 7.1\%, 4.9\%, 2.6\%, and 3.7\%, respectively.
 


\subsection{Additional Findings}
In addition to presenting the results above, we conducted further quantitative and qualitative analyses regarding the targeted features to uncover additional insights and identify key areas for follow-up work.  We describe these analyses in the subsections below.

\subsubsection{Quantitative Analysis}
To further assess the individual contributions of the targeted features, we performed a follow-up ablation study using our best-performing model.  We systematically retrained the model after removing one type (psycholinguistic, sentiment, or demographic) of targeted feature at a time, and report our findings in Table \ref{tab:ablation_study}.

Removing sentiment features left the model mostly unchanged in terms of AUC.  However, it produced slightly fewer true negatives and slightly more false positives.  Reducing false positives is important, particularly in light of the class imbalance; thus, the sentiment features give rise to a small but meaningful contribution to the model's overall performance.  Interestingly, it appears that the demographic and psycholinguistic features inform the model in similar and perhaps interchangeable ways: removing one group but retaining the other yields similar performance to that of a model utilizing both.  Future experiments can tease apart the contributions of individual psycholinguistic characteristics at a finer level.  Extending the psycholinguistic resources employed by our model such that they exhibit greater coverage may also result in increased performance from those features specifically.

\subsubsection{Qualitative Analysis}

In Table \ref{tab:misclassified} we present two samples misclassified by our model (one false positive, and one false negative).  We make note of a key distinction between the two: surprisingly, the false positive includes many interjections indicative of ``stalling'' behaviors, whereas the false negative is quite clear.  Neither of these is representative of other (correctly predicted) samples in their respective classes; rather, participants with dementia often exhibit more stalling or pausing behaviors, observable in text as an overuse of words such as ``uh,'' ``um,'' or ``oh.''  We speculate that our model was fooled into misclassifying these samples as a result of this style reversal.  Follow-up work incorporating stylistic features (e.g., syntactic variation or sentence structure patterns) may reduce errors of this nature.  Finally, we note that many prosodic distinctions between the two classes that pass through text mostly unnoticed may be more effectively encoded using audio features.  We plan to experiment with these as well as features from other modalities in the future, in hopes of further improving performance.

\begin{table*}[t]
\small
\centering
\renewcommand{\arraystretch}{1.2}
\begin{tabular}{lccccccccc}
\hline
\multicolumn{1}{|l}{\textbf{Approach}} & \textbf{Accuracy} & \textbf{Precision} & \multicolumn{1}{l}{\textbf{Recall}} & \textbf{F1}       & \textbf{AUC}      & \textbf{TN} & \textbf{FP} & \textbf{FN} & \multicolumn{1}{c|}{\textbf{TP}} \\ \hline

\textsc{OURS-Att-w No Psych.}                       & 0.8790          & 0.8870           & \textbf{0.9825}                 & 0.9319          & 0.9499          & 12.0         & 5.6        & 1.6         & 93.6                             \\
\textsc{OURS-Att-w No Sent.}                      & \textbf{0.8970}          & \textbf{0.9239}           & 0.9615                            & \textbf{0.9321}          & \textbf{0.9501} & 7.6         & 10.0        & 3.6         & 91.6                             \\
\textsc{OURS-Att-w No Demo.}               & 0.8908 & 0.9005  & 0.9789                            & 0.9308 & 0.9473          & 10.33        & 7.33         & 2         & 93.3  \\ 
\end{tabular}
\caption{Ablation study performed using our best-performing model (\textsc{Ours-Att-w}).}
\label{tab:ablation_study}
\end{table*}

\begin{table*}

\small
\centering
\renewcommand{\arraystretch}{1.2}
\begin{tabular}{|c|l|}
\hline
\textbf{False Positive} & \begin{tabular}[c]{@{}l@{}}Uh, oh I can oh you don't want me to memorize it.\\ Oh okay, the the little girl is asking for the cookie from the boy who is about to fall on his head\\ And she is going I guess ``shush'' or give me one\\ The mother laughs we don't think she might be on drugs because uh laughs \\ she is off someplace because the sink is running over\\ and uh it is summer outside because the window is open \\ and the grasses or the bushes look healthy.  And uh that's it.\end{tabular} \\ \hline
\textbf{False Negative} & \begin{tabular}[c]{@{}l@{}}Oh, the water is running off the sink\\ Mother is calmly drying a dish \\ The uh stool is going to fall over and the little boy is on top of it  getting in the cookie jar. \\ And the little girl is reaching for a cookie.\\ She has her hands to her her finger to her lip as if she is telling the boy not to tell.\\ The curtains seem to be waving a bit, the water is running. that's it.\end{tabular}                                                                              \\ \hline
\end{tabular}
\caption{Samples misclassified by our model. False Positives are control patients classified as AD patients, while False Negatives are AD-patients classified as control patients.}
\label{tab:misclassified}

\end{table*}

\section{Discussion}
The introduction of sentiment-based, psycholinguistic, and demographic features improved the performance of the model, demonstrating that implicitly-learned features (although impressive) still cannot encode conversational characteristics of dementia to the extent that other, more targeted features can.  Likewise, in both \textsc{C-LSTM} and our approach, the introduction of a bi-directional LSTM with an attention mechanism led to performance improvements on classifier AUC.  This improvement suggests that these additions allowed the model to better focus on specific patterns indicative of participants suffering from dementia.

In contrast, the benefits of adding class weights to the model's loss function were less clear.  We introduced this correction as a mechanism to encourage our classifier to make fewer false positive predictions, and although this worked, the model also became less capable of identifying true positives.  Given the nature of our classification problem, this trade-off is rather undesirable, and additionally this correction did not improve the general quality of the classifier---the AUC for both our model and \textsc{C-LSTM} remained almost unchanged.  However, regardless of the inclusion of class weights for the loss function, our measures regarding the AUC, Precision, Recall, F1 Score, and Accuracy show that overall our model is able to outperform the previous state of the art (\textsc{C-LSTM}) at predicting whether or not participants are suffering from dementia based on their conversational transcripts. 

\section{Conclusion}
In this work we introduced a new approach to classify conversational transcripts as belonging to individuals with or without dementia. Our contributions were as follows:
\begin{enumerate}
    \item We introduced a hybrid architecture that allowed us to take advantage of both engineered features and deep-learning techniques on conversational transcripts.
    \item We explored the effects of a bi-directional LSTM and attention mechanism on both our model and the current state of the art for dementia detection.
    \item We examined the effects of loss function modification to take into consideration the class imbalance in the DementiaBank dataset.
\end{enumerate}

Importantly, the model that we present in this work represents the new state of the art for AD detection on the DementiaBank dataset.  Our source code is available publicly online.\footnote{\url{https://github.com/flaviodipalo/AlzheimerDetection}}  In the future, we plan to explore additional psycholinguistic, sentiment-based, and stylistic features for this task, as well as to experiment with features from other modalities. Finally, we plan to work towards interpreting the neural features implicitly learned by the model, in order to understand some of the latent characteristics it captures in AD patients' conversational transcripts.

\bibliography{dipalo_parde}
\bibliographystyle{acl_natbib}

\end{document}